\documentclass{article}
\usepackage{spconf,amsmath,graphicx,hyperref,booktabs,xcolor,eso-pic}

\title{The Curious Case of Visual Grounding: \\Different Effects for Speech- and Text-based Language Encoders}
\name{Adrian Sauter \qquad Willem Zuidema \qquad Marianne de Heer Kloots}
\address{Institute for Logic, Language and Computation \\ University of Amsterdam}
\begin{document}
\AddToShipoutPictureFG*{%
  \AtPageLowerLeft{%
    \put(\dimexpr0.5\paperwidth-0.5\textwidth,30){%
      \parbox{\textwidth}{%
        \centering
        \small
        \textit{This work has been submitted to the IEEE for possible publication. \\Copyright may be transferred without notice, after which this version may no longer be accessible.}
      }%
    }%
  }%
}

\ninept
\maketitle
\begin{abstract}

How does visual information included in training affect language processing in audio- and text-based deep learning models?
We explore how such \emph{visual grounding} affects model-internal representations of words, 
    and find substantially different effects in speech- vs. 
    text-based 
    language encoders.
    Firstly, global representational comparisons reveal that visual grounding increases alignment between 
    representations of spoken and written language, but this effect seems mainly driven by enhanced encoding of word identity rather than meaning. We then apply targeted clustering analyses to probe for phonetic vs. semantic discriminability in model representations. Speech-based representations remain phonetically dominated with visual grounding, but in contrast to text-based representations, visual grounding does not improve semantic discriminability. Our findings could usefully inform the development of more efficient methods to enrich speech-based models with visually-informed semantics.
\end{abstract}
\begin{keywords}
visual grounding, self-supervised learning, \\interpretability, representational analyses, semantic structure
\end{keywords}
\section{Introduction}
Humans learn language from rich, multimodal input, making the linguistic representations they develop \emph{grounded} in multiple modalities. In contrast, 
systems for computational language understanding are traditionally designed to derive their 
representations from a single input modality, like speech or text. 
While this has been a natural starting point, multimodal learning paradigms arguably allow for richer concept representations than can be inferred from speech or text alone. 
Several lines of current research have therefore developed ways to incorporate other modalities -- typically, the visual modality -- into the training pipeline. Specifically, 
visually grounded language encoders, in both the text \cite{radford2021CLIP, zhang2021VGBert} and speech \cite{peng2022fast_vgs_plus, chrupala2022survey} domain, 
learn cross-modal language representations that are useful 
for applications bridging linguistic and visual information, including multimodal (spoken) dialogue and retrieval systems \cite{wang_crossmodalretrieval, cheng_multimodaldialogue}. 

One reported benefit of visually grounded representation spaces is their enhanced semantic structuring. Compared to their unimodal counterparts, Text-based Language Encoders (TLEs) enriched with visual information in training show higher alignment with human word similarity ratings \cite{pezzelle2021word} and better clustering by broad semantic categories \cite{zhang2021VGBert}. The semantic enhancement offered by visual grounding would seem to be particularly promising for Speech-based Language Encoder models (SLEs), given that unimodal SLE representations are generally dominated by sound- rather than meaning-based similarities \cite{choi2024main}, and capture less abstract semantic concepts than text-based models \cite{ersoy25_interspeech}. However, 
there is currently no direct comparison of the effects of visual grounding on speech- vs. text-based encoder models that would demonstrate such an advantage. 

In this paper, we therefore analyze word representations in a visually grounded SLE, and compare them to its ungrounded counterpart as well as to 
text-based models. In our first set of analyses, we find that visually grounded speech representations globally show greater alignment with text-based models, but no particular enhancement of meaning-based similarities. In follow-up analyses, we shift our focus to 
\emph{subspaces} of the embedding space tuned for specific kinds of information. 
We design two 
datasets of phonetic and semantic word groups, to measure the degree of clustering for each in both visually grounded and ungrounded models. While phonetic clustering remains similar, semantic clustering in speech-based language encoders shows surprising degradation after visual grounding. These results suggest that current approaches to visual grounding can only enhance the clustering of semantic categories in representation spaces which already saliently encode meaning — an important consideration for future developments in cross-modal representation learning. All code and data are provided for reproducibility\footnote{\url{https://github.com/adrian-sauter/visual_grounding_speech_analysis}}.

\section{Experimental setup}\label{sec:Setup}

\subsection{Extracting model representations}
We consider one pair of SLEs 
and one pair of TLEs
, obtaining pre-trained checkpoints from their publicly available sources. 
%
%
\textbf{wav2vec2} \cite{baevski2020wav2vec} is a self-supervised SLE pre-trained on LibriSpeech \cite{panayotov2015librispeech} to discriminate masked  20 ms audio frame representations generated by its convolutional feature encoder. Its Transformer module is optimized by a contrastive loss to 
distinguish positive and negative samples. 
\textbf{FaST-VGS+} \cite{peng2022fast_vgs_plus} is a visually grounded SLE, extending the pre-trained wav2vec2-base model with a visual encoder branch and fine-tuning both branches on LibriSpeech \cite{panayotov2015librispeech} and SpokenCOCO \cite{hsu2020spokencoco}, using a combination of cross-modal contrastive loss and the wav2vec2 self-supervised loss. Since its audio branch is initialized from wav2vec2-base (the ungrounded SLE in our experiments), FaST-VGS+ enables an isolated analysis of the impact of visual grounding on speech representations. 

\textbf{BERT} \cite{devlin2018bert} is a TLE, pre-trained on 3,300M words of books and web-content to capture contextualized word representations by predicting masked words in a sentence. 
\textbf{Visually-Grounded BERT} (VG-BERT) \cite{zhang2021VGBert} extends BERT by integrating a pre-trained vision encoder and fine-tuning it with a contrastive loss on MS COCO \cite{lin2014coco}.

For all models, we extract layerwise word-level representations by mean-pooling over audio frame or text token representations for single-word inputs (visual input is omitted at inference time). 

\subsection{Representational analyses}

\subsubsection{Global representational comparisons \& word pairs}
\label{subsec:cka-wordpairs}
We first ask how visual grounding globally affects the representation space of speech- and text-based models, as well as the alignment between them. We measure the similarity between model-internal representations of speech vs. text inputs using Centered Kernel Alignment (CKA; \cite{kornblith2019similarity}). CKA 
allows for comparisons between models with different architectures and representation spaces, by comparing pairwise similarity matrices 
extracted from each individual model. For this purpose, we create a large dataset of parallel SLE and TLE embeddings by sampling words from LibriSpeech \cite{panayotov2015librispeech} using existing forced alignments, yielding over 9,500 unique words and 70,000 word tokens from 80 speakers. Since LibriSpeech contains spoken sequences rather than individual words, we obtain word-level representations by audio slicing, i.e. using only the isolated target word as model input and excluding surrounding context (following \cite{choi2024main}).

We next zoom in on similarities between particular classes of word pairs to more specifically analyze what types of information drive changes in SLE word representations after visual grounding. Replicating earlier analyses on unimodal SLEs \cite{choi2024main}, we measure cosine similarities between same-speaker, same-word, near-homonym, and synonym pairs. These analyses aim to quantify the salience of different types of information (i.e. speaker identity, word identity, word pronunciation, word meaning) in model-internal representations.
We follow \cite{choi2024main} in obtaining sets of audio-sliced word pairs from LibriSpeech and additional sources of word annotations. Synonyms are obtained through the synsets (cognitive synonyms) in WordNet \cite{fellbaum2010wordnet}. To find near-homophones, we compute the normalized Levenshtein distances between phonemic transcriptions of word pairs from the CMU Pronunciation Dictionary \cite{cmu2014cmu}. Near-homophones are defined as word pairs with normalized Levenshtein distance $d \leq 0.4$, corresponding to the top 0.1\% of random word pairs in LibriSpeech. Same-word pairs (identical words) and same-speaker pairs (unique words from the same speaker) are obtained from annotations included in LibriSpeech. Finally, random word pairs form a lower bound on cosine similarities, and similarities between all other word pair classes are normalized by subtracting the mean similarity between random pairs. To assess the reliability for our word-pair analyses, we sample 10k word utterance five times, and report the average value across the five experiments with 95\% confidence intervals.

\subsubsection{Phonetic \& semantic clustering}
\label{subsec:clustering}
While previous work has shown that unimodal SLE representations are more dominated by phonetic than semantic information \cite{choi2024main, ersoy25_interspeech}, analyses involving learned projections have shown that word representations from those same SLEs do show significant alignment with text-based semantic embeddings like GloVe \cite{pasad2021layer}, and also align with sentence-level semantic similarity judgments better than acoustic and text-only baselines \cite{pasad2024words}. To investigate the \textit{decodability} of different kinds of word-level information as compared to their salience, we measure the degree of clustering for two types of word categories, 
in full model-internal representations as compared to two dimensionality-reduced subspaces. We quantify the degree of clustering by computing silhouette coefficients, which measure how well datapoints fit with their assigned groups by comparing mean cosine distances for within-group datapoints to mean distances for across-group datapoints \cite{rousseeuw1987silhouette}. We compare dimensionality-reduced projections obtained 
by Principal Component Analysis (PCA) to dimensionality-reduced projections obtained 
by Linear Discriminant Analysis (LDA). While PCA projections optimize for explained variance in the original embedding space, LDA projections are linearly optimized for separability between target groups with supervision from group labels. LDA has previously been used to analyze both audio- and text model representations \cite{deheerkloots25_interspeech, bentumSpeechRegistersDiffer2019}. 

In our experiments on phonetic and semantic clustering, we make use of the Massive Auditory Lexical Decision dataset (MALD; \cite{tucker2019mald}), which consists of more than 26,000 English word recordings from a single speaker. This allows us to create our two targeted analysis datasets for studying within-speaker structuring of phonetic and semantic information, controlling for semantic and phonetic distances respectively, as well as average concreteness scores across word groups. Concreteness scores are human ratings collected by \cite{brysbaert2014concreteness}, who define the concreteness of a word as the degree to which its referent is a perceptible entity.

For our \textbf{phonetic clustering} dataset, we identified groups of phonetically similar but semantically distinct words. We employed the following process until 14 groups across 81 unique words were found, with each group containing at least 5 words (avg. 5.79 words per group) and none containing identical words. First, a random word from the MALD dataset was selected. Next, phonetically similar words were identified based on normalised Levenshtein distances between phonemic transcriptions ($d \leq 0.529$, top 10\% in MALD), while ensuring they were dissimilar ($d > 0.529$) from words in previously identified groups. Semantically similar words were excluded by retaining only those with a cosine similarity below $0.1$ in GloVe embedding space \cite{pennington2014glove}. Finally, we only included words classified as very concrete (top 25\%) or very abstract (bottom 25\%) according to the concreteness ratings provided in MALD, ultimately resulting in 7 groups of concrete words and 7 groups of abstract words. Example groups include (1) handshake, handbrake, handbook, handmaid, handmade (avg. concreteness: 4.42), and (2) assume, astute, allure, acute, akin (avg. concreteness: 1.83).

For our \textbf{semantic clustering} dataset, we manually assembled nine groups of semantically related words (shown in Table~\ref{tab:Semantic_categories}), each containing at least 8 words (avg. 13 words per group) -- e.g., `musical instruments' (words like piano, guitar, violin) and `ethical/legal terms' (words like justice, fairness, honesty). Of these semantic groups, six are made up of concrete words (avg. concreteness rating in the top 10\%), and three are made up of abstract words (avg. concreteness rating in the bottom 25\%). Semantic groups are validated using GloVe embeddings, ensuring that all pairwise cosine similarities of words within semantic groups fall within the top 15\% of all pairwise similarities in the MALD dataset. Additionally, words within semantic groups were selected to be phonetically distinct from each other (avg. norm. Levenshtein distance $d > 0.6$).
\vspace{-0.55cm}
\begin{table}[ht]
\caption{Semantic groups with average concreteness ratings and average pairwise Levenshtein distances (± 1 std. dev.)}
\centering
\renewcommand{\arraystretch}{0.9}
\setlength{\tabcolsep}{3pt}
\footnotesize
\begin{tabular}{lcc}
\toprule
\textbf{Category (\#words)} & \textbf{Avg. Concreteness} & \textbf{Avg. Phon. Dist.}  \\
\midrule
Musical instruments (10)  & 4.91 ± 0.08 & 0.64 ± 0.11 \\
Clothing articles (19)   & 4.87 ± 0.12 & 0.63 ± 0.10 \\
Vegetables (19)          & 4.86 ± 0.16 & 0.64 ± 0.13 \\
Vehicles (14)            & 4.85 ± 0.09 & 0.68 ± 0.12 \\
Building materials (16)  & 4.78 ± 0.14 & 0.67 ± 0.12 \\
Organs (8)              & 4.65 ± 0.13 & 0.65 ± 0.11 \\
Financial terms (13)     & 2.11 ± 0.40 & 0.62 ± 0.10 \\
Emotions (10)            & 2.10 ± 0.41 & 0.70 ± 0.14 \\
Ethical/ Legal terms (8) & 1.84 ± 0.36 & 0.64 ± 0.11 \\
\bottomrule
\end{tabular}
\label{tab:Semantic_categories}
\end{table}

Given the relatively small dataset sizes for our phonetic and semantic clustering analyses, we use a leave-one-out approach to report silhouette score results, iteratively excluding one category to mitigate spurious patterns and assess robustness. We report the mean over all iterations with 95\% confidence intervals.

\section{Findings}\label{sec:Findings}
\subsection{Visual grounding bridges modalities by enhancing word identity representation}\label{sec:word_pair_similarities}
\begin{figure}[h!]
    \centering
    \includegraphics[width=\linewidth]{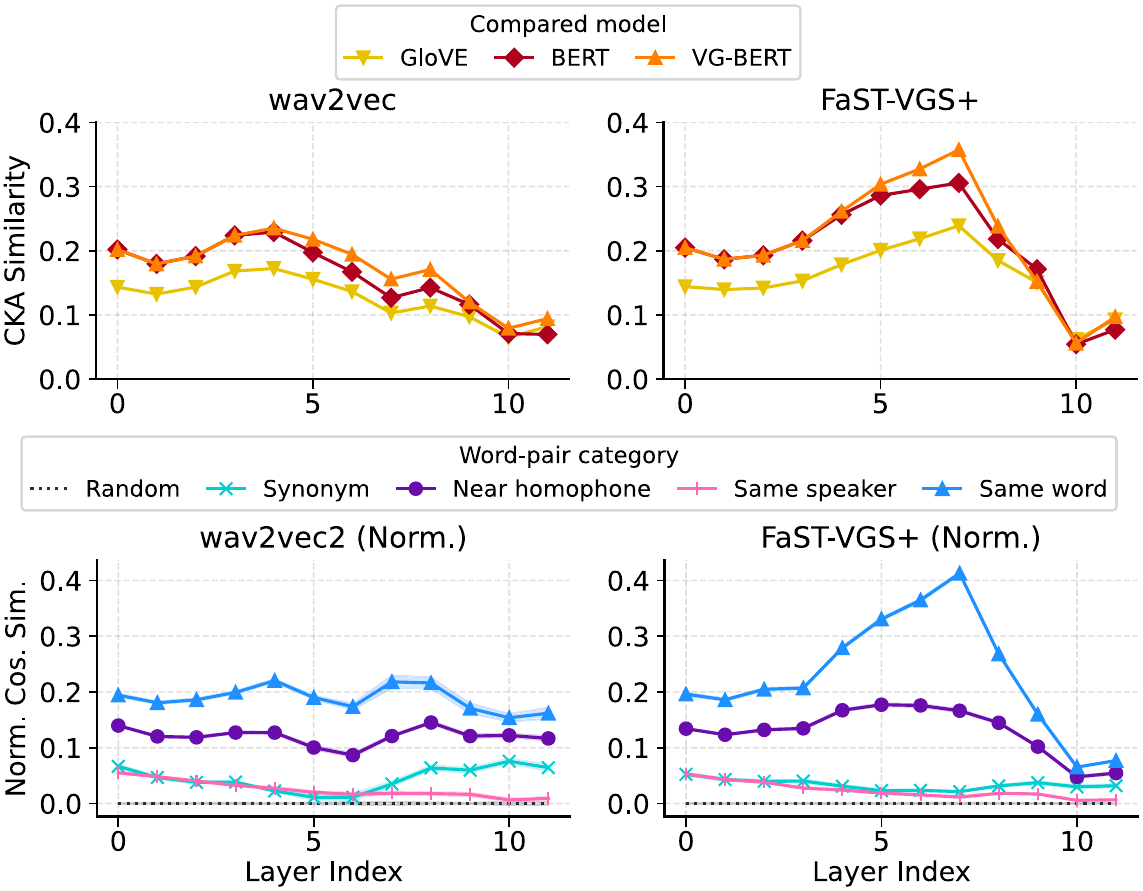} 
    \caption{Similarity between speech- and text-based models increases with visual grounding. Top plots show the layerwise CKA similarities to text-based models, bottom plots show cosine similarities between same-word pairs compared to other word pairs.}
    \label{fig:libri_speech_similarities} 
\end{figure}

Our global CKA-based comparison (Fig.~\ref{fig:libri_speech_similarities}, top row) between speech- and text-based language encoders shows generally higher similarity scores for the visually grounded \emph{FaST-VGS+} model than for its unimodal counterpart (\emph{wav2vec2}). For both SLEs, similarity is highest to \emph{VG-BERT}, followed by unimodal \emph{BERT} and static \emph{GloVe} embeddings.

Zooming in on similarities between specific word pairs in SLE representations, we find that \emph{FaST-VGS+} more saliently represents word identity than \emph{wav2vec2}, as evidenced by the strongly increased similarity scores between same-word pairs as compared to near-homophones, same-speaker pairs and synonyms (Fig.~\ref{fig:libri_speech_similarities}, bottom row). Interestingly, visual grounding does not result in more salient representations of meaning-based word information: relative similarities between synonyms are even slightly decreased in later layers of \emph{FaST-VGS+} compared to \emph{wav2vec2}. 

Taken together, these findings suggest that visual grounding improves alignment between speech- and text-based representations, but does so primarily by emphasizing word identity information rather than word meaning. This is additionally supported by the highly aligned layerwise patterns observed for between-model CKA similarities and within-model same-word similarities, both peaking in \emph{FaST-VGS+} layer 7.

\subsection{Effects of visual grounding on phonetic clustering}\label{sec:phonetic_sil_results}
\begin{figure}[h!]
    \centering
    \includegraphics[width=\linewidth]{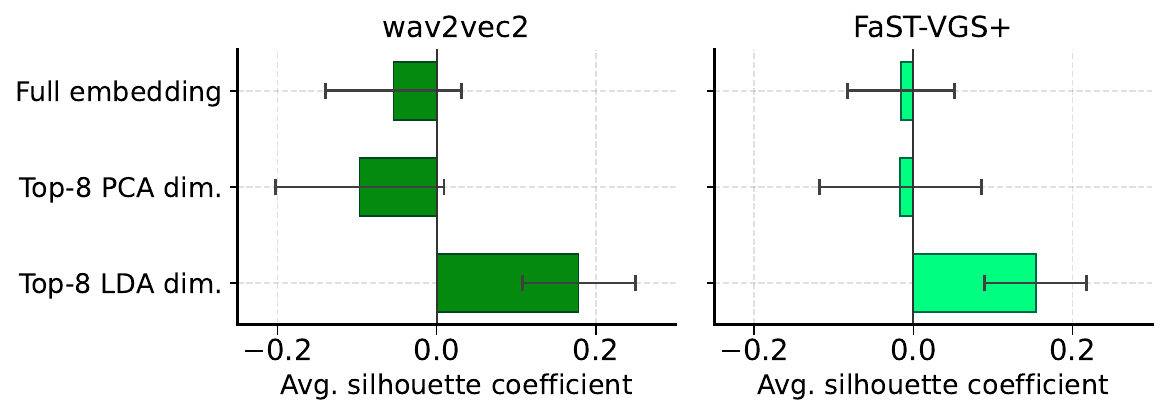} 
    \caption{Degree of clustering for predefined phonetic groups. Bars show average silhouette scores for full embeddings, PCA- \& LDA-derived subspaces. Error bars are standard deviations across layers.}
    \label{fig:phon_all_dims_vs_LDA_vs_PCA} 
\vspace{-1em}
\end{figure}
The representational analyses in the previous section revealed what information is most saliently represented in SLEs, and how this is influenced by visual grounding. As described in Section~\ref{subsec:clustering}, we will now shift our focus from representational salience to the decodability of particular categories of interest (phonetically and semantically grouped words), while avoiding speaker- and word-identity effects by analyzing groups of unique words recorded by a single speaker.


Figure~\ref{fig:phon_all_dims_vs_LDA_vs_PCA} shows the degree of clustering by phonetic groups in both SLEs, as computed across the models' full (768-dimensional) embeddings, the top-8 PCA components, and the top-8 LDA dimensions optimized for separability between phonetic groups. The full and PCA-based scores show that phonetically similar (but not identical) words are not particularly tightly clustered in either models' native embedding space, though somewhat more strongly in \emph{FaST-VGS+} than in \emph{wav2vec2} (aligning with the slightly increased similarities observed for near-homophones in Figure \ref{fig:libri_speech_similarities}). However, the LDA-based scores show that phonetic groups are successfully decodable from both SLE models after an optimized linear projection. This highlights LDA's effectiveness at decoding class-specific information from model representations, and we therefore report LDA-based clustering scores for our further analyses.

\subsection{Effects of visual grounding on semantic clustering}\label{sec:semantic_sil_results}
\begin{figure}[h!]
    \vspace{-1em}
    \centering
    \includegraphics[width=\linewidth]{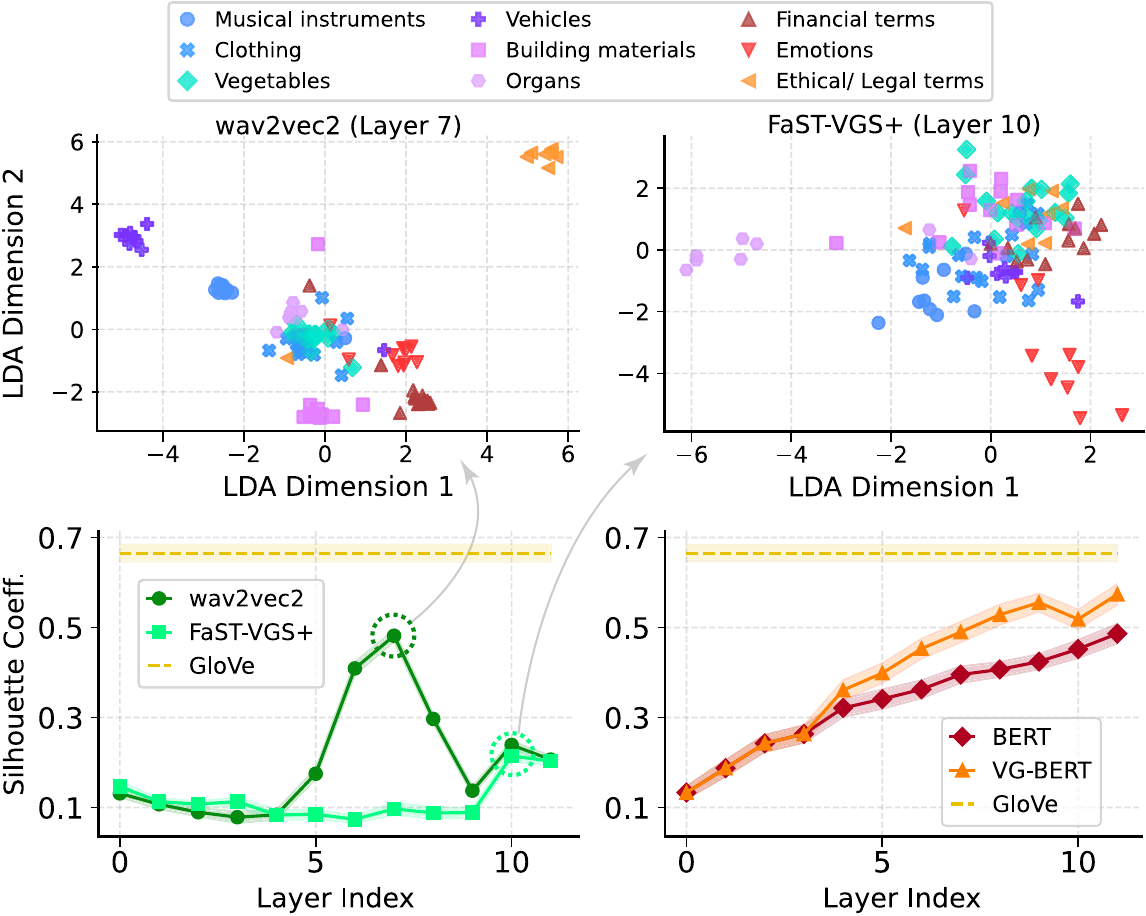}
    \caption{Visual grounding does not improve semantic clustering in SLEs, even though it does so in TLEs. Line plots show avg. silhouette scores for the top-8 LDA dimensions extracted across speech- and text-based model layers (bottom); scatter plots show the first 2 LDA dimensions for the best-performing layers of each SLE (top). }
    \label{fig:lda_plotting} 
\end{figure}

Figure \ref{fig:lda_plotting} shows the degree of clustering by semantic groups across layers of both SLE and TLE models. In line with previous results on visual grounding in TLEs \cite{zhang2021VGBert}, we find that the decodability of semantic word clusters is improved in \emph{VG-BERT} as compared to \emph{BERT} (bottom-right plot), and this effect increases from middle model layers onwards, with the final output representations of \emph{VG-BERT} nearing the topline clustering scores for \emph{GloVe} (high scores are expected for \emph{GloVe}, since we defined semantic groups based on similarity in \emph{GloVe} embedding space). In contrast, the decodability of semantic clusters in SLE representations peaks in layer 7 of the ungrounded \emph{wav2vec2} model. This peak is absent after visual grounding, with the \emph{FaST-VGS+} model also showing generally lower semantic clustering performance. The degradation of more abstract linguistic information in final model layers is a common finding in SLE representational analyses, and has been attributed to these models' low-level acoustic prediction objectives in self-supervised training \cite{pasad2021layer, deheerkloots24_interspeech}. However, the general lack of improvement in semantic clustering for \emph{FaST-VGS+} as compared to \emph{wav2vec2} reveals a striking difference in the effects of visual grounding on text- vs. speech-based word representations.

\subsection{Is word clustering affected by concreteness?}
It is a well-established finding that both unimodal and multimodal models learn concrete concepts more easily than abstract ones \cite{zhang2021VGBert, pezzelle2021word}, a trend also found in humans \cite{jessen2000concreteness}. Because we designed our word clustering datasets to include both highly concrete and highly abstract word groups (see Section~\ref{subsec:clustering}), we can examine whether this effect holds for the observed phonetic and semantic clustering patterns in SLE and TLE word representations. 

For phonetic groups, silhouette scores were indeed higher for concrete word categories as compared to abstract categories. When comparing best-layer scores, \emph{wav2vec2} (layer 7) silhouette scores were 0.29 for abstract and 0.33 for concrete categories, and \emph{FaST-VGS+} (layer 11) scores were 0.18 and 0.27 respectively.

For semantic groups, the concreteness advantage does hold in both ungrounded and grounded TLEs: average silhouette scores in \emph{BERT} layer 11 are 0.34 for abstract and 0.52 for concrete categories, compared to 0.42 and 0.62 respectively in \emph{VG-BERT} layer 11. The ungrounded \emph{wav2vec2} representations form an exception to this pattern, with higher (layer 7) scores for abstract categories (0.54) compared to concrete ones (0.50), while its visually grounded counterpart does follow the concreteness advantage despite its generally lower scores (\emph{FaST-VGS+} layer 10 representations score 0.14 on average for abstract, vs. 0.21 for concrete categories). We note that the concreteness effect is generally stronger in visually grounded models; this is likely due to the dominance of concrete concepts in the image captioning datasets used for visually grounded training.

\subsection{How do semantic subspaces change with visual grounding?}

Why does visual grounding seem to enhance semantic structure in TLEs, but have the opposite effect in SLEs? To address this question, we further analyzed the semantic LDA projections used for the clustering analyses in Section~\ref{sec:semantic_sil_results}. We now compute CKA similarities between the layerwise LDA projections from ungrounded and grounded models, to quantify the degree of semantically relevant representational changes induced by visual grounding. First, we find that grounded TLE embeddings with strong semantic clustering remain highly aligned to their ungrounded counterparts (CKA-similarity $\approx$ 0.75), while SLE embeddings change more substantially with visual grounding (CKA-similarity $\approx$ 0.46). This structural impact can also be seen in Figure \ref{fig:lda_plotting} (top row), where pre-existing semantic structuring in \emph{wav2vec2} seems to be disrupted in \emph{FaST-VGS+}. Furthermore, correlating changes in silhouette scores between grounded and ungrounded models to representational changes measured by CKA-similarities reveals a strong positive correlation for SLEs (Pearson's $r = 0.718$, $p<0.01$) but a strong negative correlation for TLEs ($r=–0.870$, $p<0.001$). Hence, SLEs layers that preserve their original geometry (thereby obtaining higher CKA-similarities) show better semantic clustering, whereas semantic clustering degrades in layers where visual grounding drives representational divergence. In contrast, TLEs achieve better semantic clustering precisely in layers that diverge most from their ungrounded form. These findings confirm that visual grounding reshapes semantically relevant dimensions of text representations to improve clustering, but disrupts equivalent dimensions in speech representations without enhancing semantic discriminability. 

\section{Discussion \& Conclusion}
We analyzed the effect of visual grounding in word representations extracted from speech-based SLEs compared to textb-based TLEs. Previous work on SLEs has measured the encoding of linguistic features using trained classification or regression probes \cite{pasad2021layer, deheerkloots24_interspeech}, or by directly analyzing distances in the representation space \cite{choi2024main, deheerkloots24_interspeech}. These two types of measures 
produce similar results when we are dealing with
features that are very saliently encoded across a model's representation space (e.g., phonetics). However, they can show intriguing differences 
when we are probing for information that is represented only in a small subspace of the whole embedding space (for instance, semantic information in an embedding space that overwhelmingly represents acoustic information).

Using global representation space comparisons and word-pair similarity analyses, we found that visual grounding aligns representations of spoken and written language, but does so mostly by enhancing word identity information. This aligns with previous findings reporting enhanced encoding of linguistic units in visually grounded SLEs \cite{peng2022fastvgs, peng22c_interspeech}, but calls into question whether visual grounding can drive similar meaning-space improvements for SLE representations as have been reported for TLEs \cite{zhang2021VGBert, pezzelle2021word}. Our clustering analyses confirm that visual grounding indeed improves semantic discriminability in TLEs, but in fact disturbs pre-existing semantic structuring in SLEs. While surprising, this finding aligns with synonym similarities being slightly more prominently encoded in ungrounded vs. grounded SLE representations, and thus proves consistent across multiple analysis techniques.

We have focussed on developing methods for targeted \emph{intrinsic} evaluations of SLE and TLE word representations, without directly examining extrinsic effects on downstream tasks. Relating differences in model-internal geometry to downstream task performance remains an important area of future work, which could further refine the significance of our findings. Moreover, we believe that our results provide relevant insights for the development of novel visually grounded training paradigms for speech-based language encoders. Existing semantic structuring in text-based model output representations affords the refinement of such structuring into more humanly intuitive distinctions, for example by optimizing visually-informed loss functions. On the other hand, semantic structuring in ungrounded speech-based models peaks in middle model layers, and may be restricted to a relatively small subspace of model-internal representations. Hence, visual grounding could potentially be more effective at semantically enhancing speech-based representations if visual information can specifically target semantically relevant dimensions present in middle model layers, rather than acoustically dominated output representations. We hope the insights and analysis datasets developed in our study will stimulate such further work into learning more effective cross-modal representations for diverse applications in speech- and language technology.

\newpage
\bibliographystyle{IEEEbib}
\bibliography{strings,refs}

\end{document}